# A Customer-Level Fraudulent Activity Detection Benchmark for Enhancing Machine Learning Model Research and Evaluation

Phoebe Jing, Yijing Gao, Xianlong Zeng[1]

## Abstract

In the field of fraud detection, the availability of comprehensive and privacy-compliant datasets is crucial for advancing machine learning research and developing effective anti-fraud systems. Traditional datasets often focus on transaction-level information, which, while useful, overlooks the broader context of customer behavior patterns that are essential for detecting sophisticated fraud schemes. The scarcity of such data, primarily due to privacy concerns, significantly hampers the development and testing of predictive models that can operate effectively at the customer level. Addressing this gap, our study introduces a benchmark that contains structured datasets specifically designed for customer-level fraud detection. The benchmark not only adheres to strict privacy guidelines to ensure user confidentiality but also provides a rich source of information by encapsulating customer-centric features. We have developed the benchmark that allows for the comprehensive evaluation of various machine learning models, facilitating a deeper understanding of their strengths and weaknesses in predicting fraudulent activities. Our dataset includes detailed attributes that reflect customer behavior over time, thereby enabling researchers to employ and test advanced analytical techniques, such as deep learning and anomaly detection algorithms, under realistic scenarios. By offering the dataset to the research community, we aim to set a new standard in fraud detection research, providing a tool that can significantly enhance the predictive accuracy of fraud detection systems. This initiative not only fosters innovation in machine learning model development but also contributes to safer banking practices, ultimately protecting consumers and financial institutions alike from the perils of fraudulent activities. Through this work, we seek to bridge the existing gap in data availability, offering researchers and practitioners a valuable resource that empowers the development of next-generation fraud detection techniques. Our code and data will be open-source and publicly available at Github.

## Introduction

Fraud detection remains a critical challenge in the banking sector, where the ability to quickly and accurately identify fraudulent activities can significantly influence the financial stability of institutions and the security of customer assets. Machine learning (ML) has emerged as a pivotal tool in combating fraud, leveraging vast amounts of data to discern patterns and anomalies that human analysts might miss. The efficiency of ML in fraud detection is largely contingent upon the quality and scope of the data used. However, while these systems are potent in theory, their practical efficacy is often hindered by the limitations inherent in the available datasets.

[1] xz926813@ohio.edu

Most existing datasets for fraud detection are primarily focused on transaction-level data. These datasets typically provide detailed information about individual transactions but lack broader behavioral contexts that might provide insights into customer-level patterns of fraudulent behavior. This focus can lead to significant gaps in understanding and predicting fraud because it fails to capture the cumulative anomalies or consistent patterns of behavior at the customer level, which are often more indicative of fraud. Furthermore, the availability of these datasets is significantly constrained by privacy concerns. Banks and financial institutions must adhere to strict data protection regulations, such as the GDPR in Europe and various other privacy frameworks globally, which limit the sharing and utilization of sensitive customer data for research purposes.

To address these challenges, our study proposes the development of a structured, customer-level dataset designed specifically to overcome these limitations while ensuring compliance with privacy regulations. This new dataset aims to enrich the resources available to the research community, providing a more comprehensive view of customer behavior and facilitating the development of more sophisticated and accurate ML models for fraud detection. By focusing on the aggregation of customer-level data, our approach offers a novel perspective in the study of financial fraud, paving the way for breakthroughs in machine learning techniques and their applications in real-world scenarios.

The contribution of our study is outlined as follows:
- Development of a Customer-Level Benchmark. Our study addresses the critical gap in available data for fraud detection by creating a structured customer-level dataset. The dataset includes detailed behavioral patterns and transactional histories at the individual customer level, rather than just isolated transaction events. This broader data scope allows for more accurate modeling of potential fraud.
- Benchmark Creation for Machine Learning Models. By providing a benchmark with the created new dataset, our study facilitates a standardized way for researchers and practitioners to evaluate and compare the performance of various machine learning models in detecting fraud. This benchmark includes performance metrics that are relevant to both the accuracy and recall/precision of fraud detection, crucial for practical implementations.
- Supporting the Research Community and Enhancing Collaborative Efforts. By making the dataset publicly available to the research community, Our study encourages collaborative efforts and cross-institutional research, fostering innovation and improvement in fraud detection technologies. This contribution is vital in a field where shared knowledge and resources can lead to significantly enhanced solutions.

# Benchmark

This section introduces our newly created Customer-level Fraud Detection Benchmark (CFDB). Initially, we will describe the two primary data sources utilized to construct this customer-centric benchmark: the SAML-D dataset and the AML-World dataset, specifically focusing on its HI-Small and LI-Small subsets. Following the overview of these foundational datasets, we will delve into the statistics of our transformed benchmark, highlighting the key metrics and insights derived from converting transaction-level data into aggregated customer profiles. This structured approach not only demonstrates the methodology behind the CFDB creation but also underscores the enhanced capability of this benchmark to facilitate more effective and nuanced fraud detection analyses.

## AMLworld

The AML-World dataset [2], developed by IBM, offers a unique and sophisticated synthetic environment for advancing Anti-Money Laundering (AML) research[2]. Designed to emulate a comprehensive financial ecosystem, this dataset captures interactions among individuals, companies, and banks, reflecting the complexity of real-world financial networks. The virtual setting provides a canvas for a wide range of financial activities such as consumer purchases, industrial procurements, salary payments, and more, making it a rich resource for developing and testing AML models. Distinctly, the AML-World dataset includes both high illicit (HI) and low illicit (LI) transaction groups, further segmented into small, medium, and large sets to cater to different computational and research needs. This stratification allows researchers to tailor their analysis based on the scale of data they can manage, with each dataset offering a unique set of challenges in terms of laundering detection. With over 200 million transactions detailed across these groups, and specific laundering patterns identified, the dataset serves as a critical tool for understanding and detecting complex laundering activities, including the three stages of money laundering: placement, layering, and integration. Moreover, IBM's generator tracks funds derived from illicit activities across numerous transactions, enabling precise labeling of laundering versus legitimate transactions—a capability not typically feasible with real-world data due to visibility constraints across institutions. This comprehensive view supports the creation of more effective laundering detection models that can operate on a broad scale yet apply to specific banking scenarios.

## SAML-D

The SAML-D dataset [1] represents a significant advancement in the field of Anti-Money Laundering (AML) research, crafted to address the pressing need for improved transaction monitoring systems capable of handling the complexities of modern financial crimes. Developed through a novel AML transaction generator, this dataset encompasses 9,504,852 transactions, meticulously annotated to distinguish between 11 normal and 17 suspicious typologies, reflecting a mere 0.1039% of transactions as suspicious. This rich dataset integrates a comprehensive set of 12 features, including transaction timing, detailed account information for both senders and receivers, transaction amounts, and diverse payment types such as credit cards and cross-border transfers. Furthermore, it enhances the research utility by incorporating 15 distinct graphical network structures that represent transaction flows within these typologies, adding an additional layer of complexity to challenge and refine detection algorithms. Such detailed and structured data provision is vital for researchers and practitioners aiming to develop more effective AML tools, thereby contributing significantly to the robustness of financial security systems globally. For more detailed insights and to respect intellectual property rights, users of the SAML-D dataset are encouraged to refer to its foundational paper by Oztas et al. (2023), or visit the corresponding Kaggle page[3].

---
[2] https://www.kaggle.com/datasets/ealtman2019/ibm-transactions-for-anti-money-laundering-aml

## Customer-level Fraud Detection Benchmark

The Customer-level Fraud Detection Benchmark (CFDB) represents a transformative approach in fraud detection research by converting transaction-level datasets into a customer-centric framework. This benchmark incorporates three pivotal datasets: SAML-D, AML-World-HI-Small, and AML-World-LI-Small, all restructured to emphasize customer behavior over individual transactions. This shift is crucial as it aligns more closely with the operational realities of financial institutions, where understanding the broader context of a customer's financial activities is often more indicative of fraudulent behavior than isolated transaction analysis.

In detail, as shown in table 1, the conversion process aggregates transactions into customer profiles, focusing on the total number and nature of transactions per customer, average transaction amounts, and the incidence of flagged suspicious activities. For instance, the CFDB now includes 855,460 customers in the SAML-D dataset, with 0.92% flagged as suspicious. Similarly, the AML-World-HI-Small and AML-World-LI-Small datasets contain 705,907 and 515,088 customers, respectively, with 0.751% and 1.23% flagged as suspicious. This aggregation allows for a more nuanced analysis of patterns that might indicate fraudulent activities, such as unusually high transaction volumes or amounts relative to the typical customer profile within the dataset.

Furthermore, the CFDB enriches the field of fraud detection by providing a dataset that is not only large-scale but also varied in terms of transaction types and customer behaviors. For example, the average transaction count per customer ranges from 9.80 to 11.11 across the datasets, with average transaction amounts showing substantial variance, illustrating different levels of financial activity and potential risk exposure. By offering these detailed insights, the CFDB enables researchers and practitioners to develop, test, and refine advanced machine learning models tailored to detect and understand complex fraudulent schemes at the customer level, significantly enhancing the predictive capabilities and operational effectiveness of anti-fraud systems.

---

[3] https://www.kaggle.com/datasets/berkanoztas/synthetic-transaction-monitoring-dataset-aml/data

Table 1. Customer-Level Statistics for the Customer-level Fraud Detection Benchmark

|  | SAML-D | AML-World-HI-Small | AML-World-LI-Small |
|---|---|---|---|
| #Customer | 855,460 | 705,907 | 515,088 |
| #Alerted Customer | 7,902 | 5,304 | 6,357 |
| %Alerted Customer | 0.92 | 0.751 | 1.23 |
| #Transactions | 9,504,852 | 6,924,049 | 5,078,345 |
| Avg. Transaction Count | 11.11 | 9.80 | 9.85 |
| Avg. Transaction Amount | 8,762 | 5,500,051 | 5,248,999 |

# Experiment Setup

In this section, we outline the experimental framework established to evaluate the performance of various machine learning models using the Customer-level Fraud Detection Benchmark (CFDB). The experiment involves a systematic analysis of the models' ability to accurately identify fraudulent activities based on the customer profiles derived from the SAML-D, AML-World-HI-Small, and AML-World-LI-Small datasets. We detail the preprocessing steps, such as normalization of data and handling of imbalanced classes, followed by the specification of training, validation, and testing splits. The objective is to ensure that each model is assessed under uniform conditions, allowing for fair comparison and robust validation of their predictive capabilities.

## Baselines

For the purpose of benchmarking the CFDB, we establish several baseline models, each representing a different approach in machine learning, to evaluate their efficacy in detecting fraudulent behavior at the customer level. These models were chosen based on their prevalence in the literature and their diverse methodologies, ranging from simple to complex algorithms. Here, we describe the setup and rationale for each chosen baseline model.

***Linear Regression.*** Linear Regression serves as our initial baseline model. Despite its simplicity and assumption of linearity between the input features and the target variable, it provides a useful benchmark for performance comparison. This model helps in understanding the direct relationships between features and the likelihood of fraud, acting as a baseline for evaluating whether more complex models significantly outperform simpler approaches in this specific context.

***Decision Tree.*** The Decision Tree model is selected for its interpretability and ability to handle nonlinear relationships. It works by creating a tree-like model of decisions, making it particularly useful for understanding the decision rules that can indicate fraudulent behavior. This model will be evaluated for its depth, the complexity of the decisions, and its robustness against overfitting, especially given the potentially noisy and complex nature of financial transaction data.

***XGBoost.*** XGBoost [17] is a highly efficient and scalable implementation of gradient boosting, is chosen for its strong performance in classification tasks involving structured data. Known for its speed and performance, XGBoost will be assessed for its ability to handle imbalanced data, its parameter tuning, and its effectiveness in enhancing predictive accuracy over more traditional methods.

***Neural Network.*** Neural Network model is included to capture complex patterns and interactions in the data that may not be readily apparent or effectively modeled by simpler algorithms. Specifically, deep learning models, such as RNN, CNN, Transformer[12-14] have demonstrate its impressive modeling power on many domains, such as natural language processing[16], healthcare[3-10], image recognition[15], and many others. Given the high dimensionality and potential non-linear relationships in the CFDB, Neural Networks could provide significant improvements in detecting subtle signs of fraudulent activities. In this study, we employ a three-layer feedforward architecture to handle the complexities in the data. The structure includes an input layer, one hidden layer, and an output layer, optimize the model's performance in detecting fraudulent activities within customer-level data.

## Implementation Details

In the implementation phase of our experiments with the Customer-level Fraud Detection Benchmark (CFDB), we utilized Python programming language along with libraries such as Pandas for data manipulation, Scikit-Learn for machine learning model implementation, and XGBoost and TensorFlow for more advanced model architectures. The preprocessing of data involved handling missing values, encoding categorical variables, and scaling numerical features to ensure uniformity across different models. The environment was set up to be reproducible, with fixed seeds for random number generators where applicable, to ensure that the results are consistent and can be validated independently.

## Training Details

The training of the models was conducted on a Mac mini (M1, 2020) to accommodate the computational demands of large-scale data processing and model training. Each model was trained using a stratified split of 70% of the data for training and 30% for testing, ensuring that the distribution of classes was consistent across all splits. Hyperparameters for each model were used default parameter. No special attention was given to handling the imbalanced dataset typical of fraud detection tasks. Methods such as SMOTE for oversampling the minority class is a limitation of the study, and will keep as a future work.

## Evaluation Metrics

To assess the performance of each model on the CFDB, we employed a variety of evaluation metrics that provide a comprehensive view of each model's effectiveness in detecting fraudulent transactions:
- **Precision**: Measures the accuracy of positive predictions, i.e., the proportion of predicted fraudulent transactions that were actually fraudulent. This metric is crucial in the context of fraud detection where the cost of false positives can be significant.
- **Recall**: Also known as sensitivity, it measures the ability of the model to detect all relevant instances, i.e., the proportion of actual fraudulent transactions that were correctly identified by the model. High recall is critical in minimizing the risk of missed frauds.
- **Accuracy**: Provides a general indication of the model's ability to correctly label both fraudulent and non-fraudulent transactions. However, due to the imbalanced nature of fraud detection datasets, this metric can be misleading if used in isolation.
- **AUC (Area Under the Curve)**: Represents the area under the ROC curve and provides an aggregate measure of performance across all possible classification thresholds. A higher AUC indicates a better performing model, with greater distinguishability between positive and negative classes.
- **F1 Score**: The harmonic mean of precision and recall, providing a single score that balances both the concerns of precision and recall in one number. It is particularly useful when dealing with imbalanced datasets.

## Result and Discussion

As shown in Figure 1. The evaluation of various machine learning models on the SAML-D dataset within the Customer-level Fraud Detection Benchmark (CFDB) demonstrates a range of performances across different metrics. Each model was assessed based on its accuracy, precision, recall, AUC, and F1 score.

Linear Regression achieved high overall accuracy (99%) but was less effective in precision (0.63) and particularly recall (0.08), indicating a significant number of false negatives. The F1 score, which balances precision and recall, was notably low at 0.14, reflecting the model's poor performance in correctly identifying fraudulent cases despite its overall accuracy. Decision Tree models displayed perfect accuracy and a strong balance between precision (0.77) and recall (0.74), resulting in a robust F1 score of 0.75. This model's performance suggests it is highly effective at identifying fraudulent transactions with fewer false positives and negatives, although its AUC was the lowest among the models at 0.87. XGBoost showed outstanding performance across all metrics, achieving perfect scores in accuracy, AUC (1.0), and an excellent balance with an F1 score of 0.85. Its precision (0.85) and recall (0.84) were also superior, indicating a high degree of reliability in identifying both positive and negative classes. Neural Network also demonstrated high effectiveness with an accuracy of 99.62%, and very competitive precision (0.81) and recall (0.78). The AUC was high at 0.9816, and the F1 score was strong at 0.801, suggesting good overall performance but slightly trailing behind XGBoost in terms of precision and recall balance.

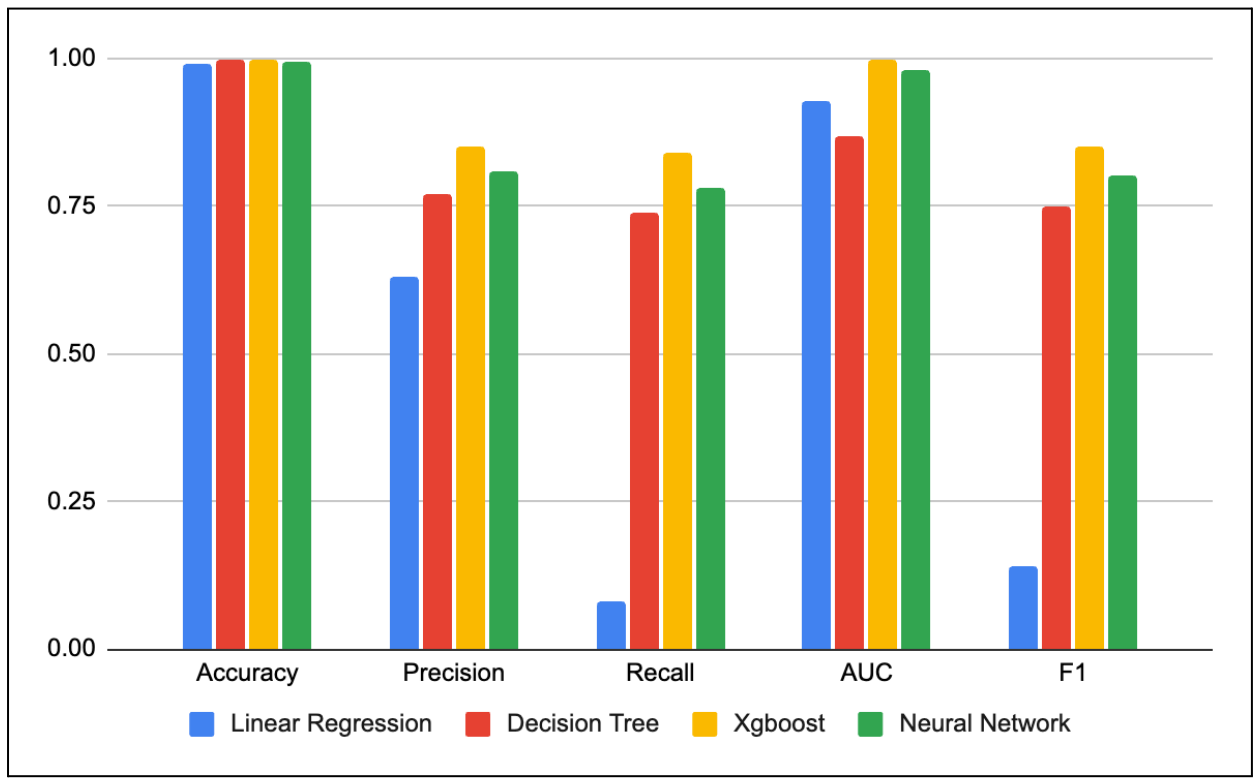

Figure 1. Comparative Performance of Machine Learning Models on the SAML-D Data Source

The performance of the models on the AML-World LI-Small and HI-Small datasets further underscores the complexities of fraud detection across different data contexts, as shown in Figure 2 and Figure 3. Here, we analyze how each model fares on datasets with lower and higher illicit transaction ratios. For model performance on AML-World LI-Small. Linear Regression showed high accuracy (99%) but completely failed in recall (0%), indicating it did not correctly identify any fraudulent transactions. The precision was perfect (1), though misleading as it did not truly capture any fraud, reflected in an F1 score of 0.01. Decision Tree presented a slightly better balance but still underperformed with low precision (0.04) and recall (0.05), resulting in an F1 score of 0.04. The low AUC (0.52) highlights its inability to differentiate between the classes effectively. XGBoost achieved decent accuracy (99%) and precision (0.74), but like linear regression, it struggled significantly with recall (0.02), leading to an F1 score of 0.04. This suggests difficulty in capturing the sparse instances of fraud in the dataset. Neural Network mirrored the decision tree's performance with low precision (0.03) and recall (0.01), culminating in a similarly low F1 score (0.03). The AUC was marginally higher at 0.71.

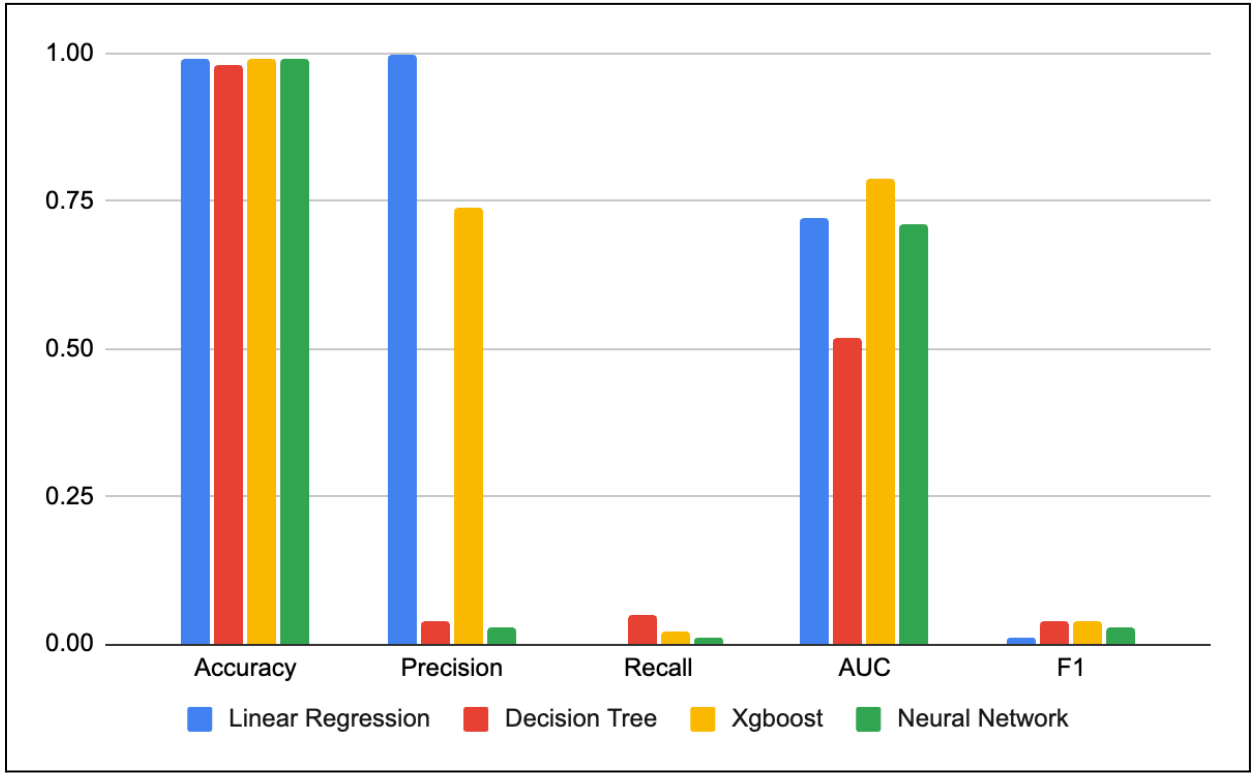

Figure 2. Comparative Performance of Machine Learning Models on the AML-World LI-Small

For model performance on AML-World HI-Small. Linear Regression recorded high accuracy (99%) but low precision (0.52) and even lower recall (0.02), resulting in an F1 score of 0.03. This model was ineffective at identifying the relatively higher proportion of fraudulent transactions. Decision Tree improved slightly with precision (0.14) and recall (0.16), achieving an F1 score of 0.15. The model's moderate AUC (0.58) suggests limited capability in distinguishing between classes. XGBoost continued to perform well with the highest AUC (0.82) in this group, decent precision (0.77), and modest recall (0.11), yielding an F1 score of 0.20. This indicates some success in handling complex patterns of fraud. Neural Network showed reasonable precision (0.62) but very low recall (0.01), leading to an F1 score similar to XGBoost's at 0.19. The AUC of 0.74 suggests it can somewhat differentiate fraudulent from non-fraudulent transactions.

The collective model performance across the diverse datasets of the Customer-level Fraud Detection Benchmark—SAML-D, AML-World LI-Small, and AML-World HI-Small—reveals insightful patterns and challenges inherent to the application of machine learning in the domain of fraud detection. XGBoost emerged as the standout model, consistently demonstrating strong capabilities in handling both the scarcity and complexity of fraudulent transactions across datasets. Its success underscores the advantages of sophisticated ensemble methods that are adept at discerning intricate patterns indicative of fraud.

The neural network, while presenting promising results, particularly on the SAML-D dataset, exhibited variability in its effectiveness, indicating that deep learning's potential might be contingent on dataset characteristics and the specific nature of financial transactions. Conversely, simpler models like linear regression and decision trees, despite their computational efficiency and interpretability, struggled with the critical task of balancing precision and recall—a balance essential for minimizing false positives and negatives in fraud detection.

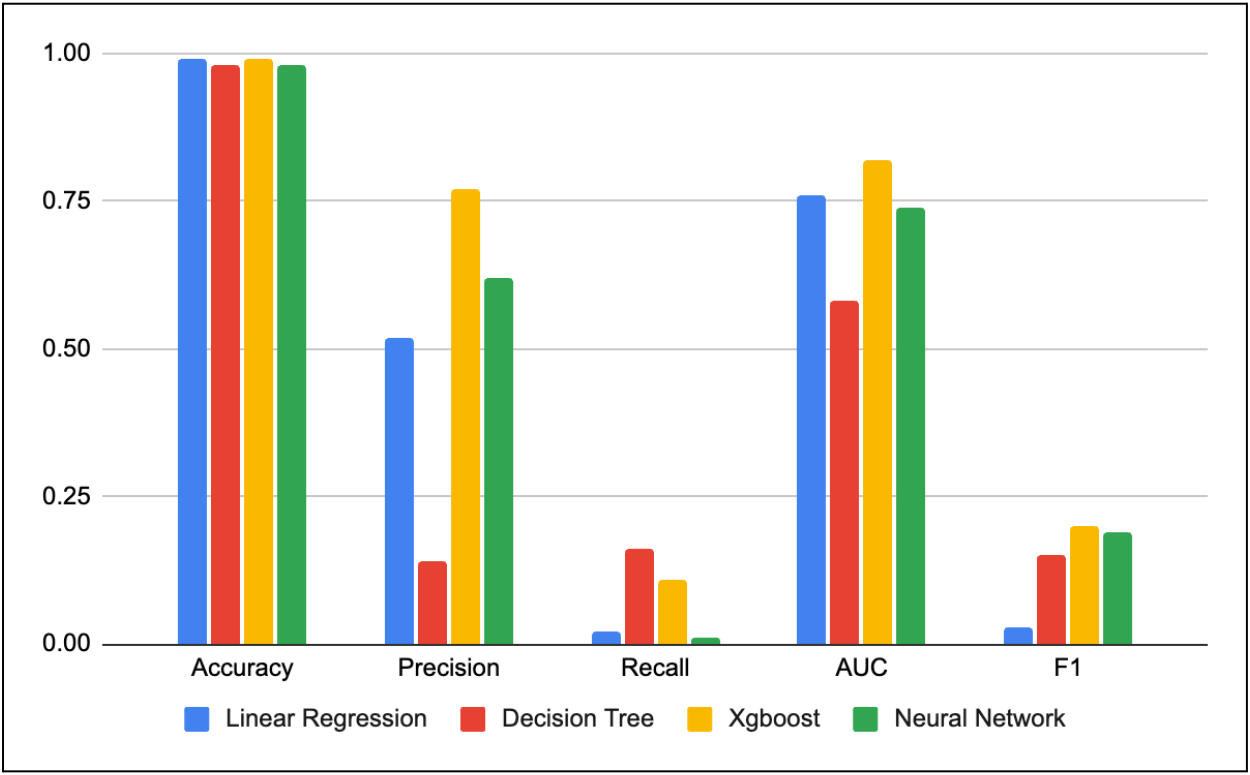

Figure 3. Comparative Performance of Machine Learning Models on the AML-World HI-Small Dataset

A significant observation across all datasets was the discrepancy between high accuracy and the lower precision, recall, and F1 scores. This discrepancy draws attention to the pitfalls of using accuracy as a sole metric in the evaluation of models, especially in imbalanced datasets typical of fraud detection. Accuracy alone fails to reflect the true efficacy of a model in identifying fraud, as evidenced by the poor recall rates in the AML-World LI-Small dataset, pointing to the models' difficulties in detecting rarer fraudulent transactions.

The findings also suggest that the complexity introduced by a higher volume of fraudulent transactions in the AML-World HI-Small dataset posed additional challenges, affecting the models' ability to distinguish between legitimate and fraudulent activity accurately. This observation signals the need for ongoing development and refinement of machine learning techniques to improve fraud detection models' sensitivity and precision.

These insights collectively highlight the necessity of a multifaceted approach in selecting and developing machine learning models for fraud detection. Rather than relying on a singular model type, there appears to be a compelling case for exploring hybrid approaches that integrate the strengths of various models to address different aspects of fraud detection. Moreover, the results emphasize the importance of developing strategies to better address data imbalances, suggesting that future research may benefit from exploring novel sampling methods, data augmentation techniques, or adaptive loss functions. Continuous evaluation and model updating are also essential, considering the dynamic and adaptive nature of fraudulent behaviors, to ensure that fraud detection systems remain effective over time.

# Conclusion

In conclusion, our study has made considerable strides in advancing the field of fraud detection by introducing the Customer-level Fraud Detection Benchmark (CFDB), a comprehensive and privacy-compliant dataset aimed at enhancing the development and evaluation of machine learning models. Through the meticulous aggregation of transaction-level data into customer profiles, the CFDB provides an invaluable resource for capturing complex behavioral patterns indicative of fraudulent activities. Our extensive analysis across three datasets—SAML-D, AML-World LI-Small, and AML-World HI-Small—has revealed the nuanced capabilities and limitations of various machine learning models in detecting fraud. The consistently strong performance of XGBoost across all datasets underscores the effectiveness of gradient boosting methods in managing the intricate nature of financial transactions. The varying results of neural networks, while promising, point to the need for careful consideration of model architecture and dataset characteristics. The study further illuminates the challenges posed by imbalanced datasets, a common issue in fraud detection, where traditional accuracy metrics fall short. The disparity between accuracy and other critical evaluation metrics such as precision, recall, AUC, and F1 score emphasizes the importance of using a multifaceted metric approach to assess model performance accurately. As we look towards the future, the CFDB stands as a testament to the potential of machine learning in revolutionizing fraud detection. However, the journey does not end here. Ongoing research must focus on the refinement of models and the exploration of innovative approaches, such as ensemble and hybrid models, to enhance sensitivity and specificity in fraud detection. Adaptability and continuous improvement will be key as we strive to keep pace with the ever-evolving tactics of financial fraudsters. By contributing a structured, customer-level dataset and a robust benchmarking framework, this study not only aids in the immediate improvement of fraud detection methods but also sets the stage for future innovations in the field, fostering a safer financial environment for all stakeholders involved.